\documentclass[runningheads]{llncs}
\usepackage{graphicx} 
\usepackage[colorlinks=true,allcolors=blue]{hyperref}

\usepackage{array}
\newcolumntype{L}[1]{>{\raggedright\let\newline\\\arraybackslash\hspace{0pt}}m{#1}}
\newcolumntype{C}[1]{>{\centering\let\newline\\\arraybackslash\hspace{0pt}}m{#1}}
\newcolumntype{R}[1]{>{\raggedleft\let\newline\\\arraybackslash\hspace{0pt}}m{#1}}

\begin{document}
\title{Spherical U-Net on Cortical Surfaces: Methods and Applications}
%
%\titlerunning{Abbreviated paper title}
% If the paper title is too long for the running head, you can set
% an abbreviated paper title here
%
\author{Fenqiang Zhao\inst{1,2} \and
Shunren Xia\inst{1} \and
Zhengwang Wu\inst{2} \and
Dingna Duan\inst{1,2} \and
Li Wang\inst{2} \and
Weili Lin\inst{2} \and
John H Gilmore\inst{3} \and
Dinggang Shen\inst{2} \and
Gang Li\inst{2}   }
\authorrunning{F. Zhao et al.}
% First names are abbreviated in the running head.
% If there are more than two authors, 'et al.' is used.
%
\institute{Key Laboratory of Biomedical Engineering of Ministry of Education, Zhejiang University, Hangzhou, China \and
Department of Radiology and BRIC, University of North Carolina at Chapel Hill, Chapel Hill, NC, USA \\
\email{gang\_li@med.unc.edu} \and
Department of Psychiatry, University of North Carolina at Chapel Hill, \\
 Chapel Hill, NC, USA}
\maketitle              % typeset the header of the contribution
\begin{abstract}
Convolutional Neural Networks (CNNs) have been providing the state-of-the-art performance for learning-related problems involving 2D/3D images in Euclidean space. However, unlike in the Euclidean space, the shapes of many structures in medical imaging have a spherical topology in a manifold space, e.g., brain cortical or subcortical surfaces represented by triangular meshes, with large inter-subject and intra-subject variations in vertex number and local connectivity. Hence, there is no consistent neighborhood definition and thus no straightforward convolution/transposed convolution operations for cortical/subcortical surface data. In this paper, by leveraging the regular and consistent geometric structure of the resampled cortical surface mapped onto the spherical space, we propose a novel convolution filter analogous to the standard convolution on the image grid. Accordingly, we develop corresponding operations for convolution, pooling, and transposed convolution for spherical surface data and thus construct spherical CNNs. Specifically, we propose the Spherical U-Net architecture by replacing all operations in the standard U-Net with their spherical operation counterparts. We then apply the Spherical U-Net to two challenging and neuroscientifically important tasks in infant brains: cortical surface parcellation and cortical attribute map development prediction. Both applications demonstrate the competitive performance in the accuracy, computational efficiency, and effectiveness of our proposed Spherical U-Net, in comparison with the state-of-the-art methods.

\keywords{Spherical U-Net  \and Convolutional Neural Network \and Cortical Surface \and Parcellation \and Prediction.}
\end{abstract}
\section{Introduction}
Convolutional Neural Networks (CNNs) based deep learning methods have been providing the state-of-the-art performance for a variety of tasks in computer vision and biomedical image analysis in the last few years, e.g., image classification \cite{p1}, segmentation \cite{p2}, detection and tracking \cite{p3}, benefiting from their powerful abilities in feature learning. In biomedical image analysis, U-Net and its variants have become one of the most popular and powerful architectures for image segmentation, synthesis, prediction, and registration owing to its strong ability to jointly capture localization and contextual information \cite{p4}. 

However, these CNN methods are mainly developed for 2D/3D images in Euclidean space, while there is still a significant demand for models that can deal with data representation on non-Euclidean space. For example, the shapes of many structures in medical imaging have an inherent spherical topology in a manifold space, e.g., brain cortical or subcortical surfaces represented by triangular meshes, which typically have large inter-subject and intra-subject variations in vertex number and local connectivity. Hence, unlike in the Euclidean space, there is no consistent and regular neighborhood definition and thus no straightforward convolution/transposed convolution and pooling operations for cortical/subcortical surface data. Therefore, despite many advantages of CNN in 2D/3D images, the conventional CNN cannot be directly applicable to cortical/subcortical surface data. 

To address this issue, two main strategies have been proposed to extend the conventional convolution operation to the surface meshes \cite{p5}, including (1) performing convolution in non-spatial domains, e.g., the spectral domain obtained by the graph Laplacian \cite{p6,p7,p8}; (2) projecting the original surface data onto a certain intrinsic space, e.g., the tangent space (which is an Euclidean space with consistent neighborhood definition \cite{p9,p10,p11}). On one hand, recent advances in convolution in non-spatial domains \cite{p7,p8} are mainly focusing on omnidirectional image data, which is typically parameterized by spherical coordinates $ \alpha \in [0,2\pi)$ and $\beta \in [0,\pi]$. While cortical/subcortical surface data are typically represented by triangular meshes, these methods still cannot be applicable, unless the surface is resampled to obtain another sphere manifold parameterized by $\alpha$ and $\beta$. This resampling process from the spherical surface with uniform vertices to another imbalanced sphere manifold with extremely non-uniform nodes is essentially hazardous and unnecessary for cortical surface data, because it would miss key structural and connectivity information, thus leading to inaccurate and ambiguous results. On the other hand, for cortical surface data analysis, existing researches typically adopting the second strategy also suffer from some inherent drawbacks. For example, the method in \cite{p9} first projected intrinsic spherical surface patches into tangent spaces to form 2D image patches, and then the conventional CNN was applied for classifying each vertex to derive the surface parcellation map. Seong et al. \cite{p10} designed a rectangular filter on the tangent plane of the spherical surface for sex classification. They resampled points in the rectangular patches for convolution operation. For a better comparison with our proposed method, we redraw their rectangular patch (RePa) convolution method in the bottom row of Fig. \ref{fig2}A. Overall, as in \cite{p9,p10}, this projection strategy would inevitably introduce feature distortion and re-interpolation, thus complicating the network, increasing computational burden and decreasing accuracy.

\begin{figure}[t]
	\includegraphics[width=0.95\textwidth]{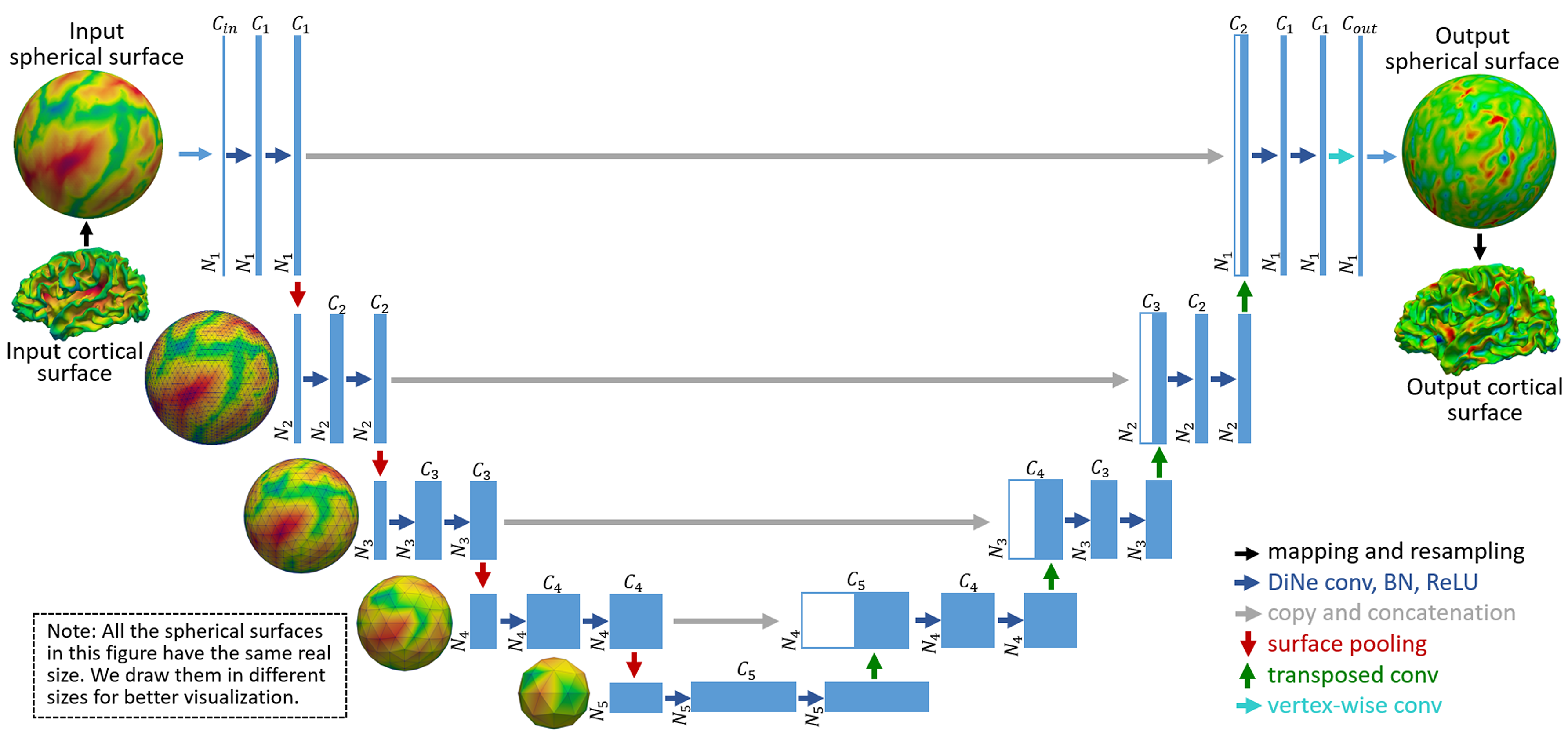}
	\caption{Spherical U-Net architecture. Blue boxes represent feature maps on spherical space. The number of features $C_i$  is denoted above the box. The number of vertices $N_i$ is at the lower left edge of the box. $N_{i+1}=(N_i+6)/4$, $C_{i+1}=C_i \times 2$. For example, $N_1$ can be 10,242, 40,962, or 163,842, and $C_1$ is typically set as 64. In our applications, the output surface is a cortical parcellation map or a cortical attribute map.} \label{fig1} 
\end{figure} 

To address these issues, in this paper, we capitalize on the consistent structure of the regularly-resampled brain cortical surface mapped onto a spherical space, by leveraging its inherent spherical topology. The motivation is that the standard spherical representation of a cortical surface is typically a uniform sphere structure that is generated starting from an icosahedron by hierarchically adding a new vertex to the center of each edge in each triangle \cite{p13}. Therefore, based on the consistent and regular topological structure across subjects, we suggest a novel intuitive and natural convolution filter on sphere, termed Direct Neighbor (DiNe). The definition of our DiNe filter is also consistent with the expansion and contraction process of icosahedron, in which vertices contribute to or aggregate from their direct neighbors' information at each iteration process. With this new convolution filter, we then develop surface convolution, pooling, and transposed convolution in spherical space. Accordingly, we extend the popular U-Net \cite{p4} architecture from image domains to spherical surface domains. To validate our proposed network, we demonstrate the capability and efficiency of the Spherical U-Net architecture on two challenging tasks in infant brains: cortical surface parcellation, which is a vertex-wise classification/segmentation problem, and cortical attribute map development prediction, which is a vertex-wise dense regression problem. In both tasks, our proposed Spherical U-Net achieves very competitive performance in comparison to state-of-the-art algorithms.

\section{Method}
The key of the Spherical U-Net is to define a consistent neighborhood orders on the spherical space, similar to the filter window in the 2D image space. In the following parts, we will first introduce the consistent DiNe filter in spherical space and then the spherical surface convolution, pooling, transposed convolution operations, and finally the Spherical U-Net architecture.

\subsection{Direct Neighbor Filter} 
\begin{figure}[t]
	\includegraphics[width=\textwidth]{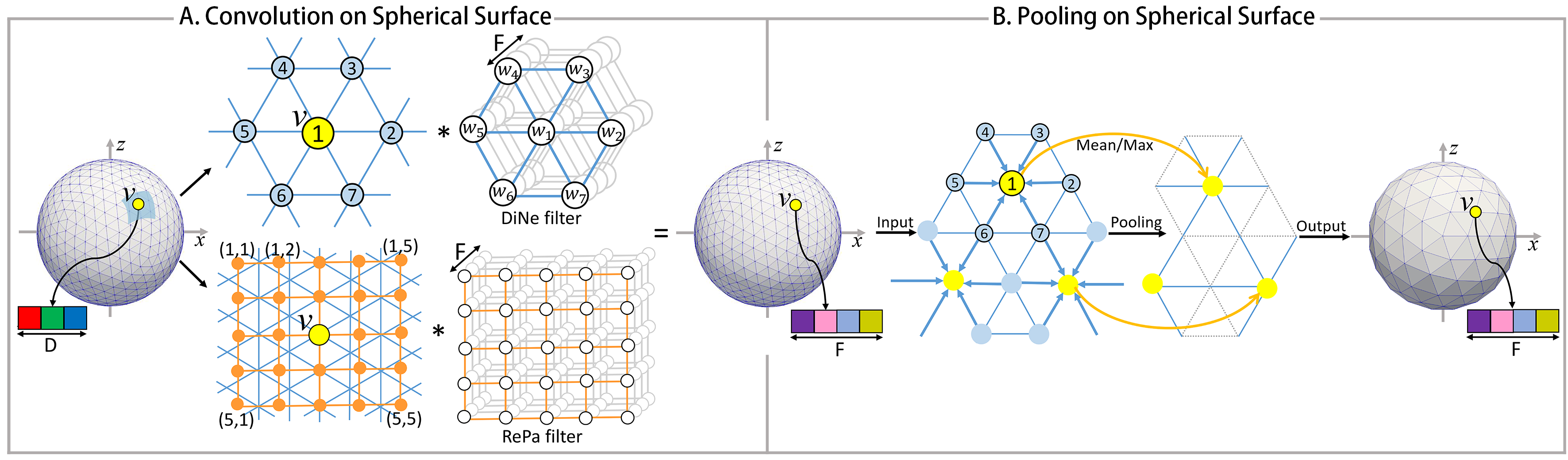}
	\caption{Top in A: Our proposed DiNe convolution. Bottom in A: The Rectangular Patch (RePa) convolution in Seong et al. \cite{p10}. Both convolutions transfer the input feature maps with $D$ channels to the output feature maps with $F$ channels. B: Illustration of the spherical surface pooling operation.} \label{fig2} 
\end{figure}

Since a standard sphere for cortical surface representation is typically generated starting from a regular icosahedron (with 12 vertices) by hierarchically adding a new vertex to the center of each edge in each triangle, the number of vertices on the sphere are increased from 12 to 42, 162, 642, 2562, 10,242, 40,962, 163,842, and so on \cite{p13}. Hence, each spherical surface is composed of two types of vertices: 1) 12 vertices with each having only 5 direct neighbors; and 2) the remaining vertices with each having 6 direct neighbors. As shown in the top row of Fig. \ref{fig2}A, for those vertices with 6 neighbors, DiNe assigns the index 1 to the center vertex and the indices 2--7 to its neighbors sequentially according to the angle between the vector of center vertex to neighboring vertex and the x-axis in the tangent plane; For the 12 vertices with only 5 neighbors, DiNe assigns the indices both 1 and 2 to the center vertex, and indices 3--7 to the neighbors in the same way as those vertices with 6 neighbors.

\subsection{Convolution and Pooling on Spherical Surface}
We name the convolution on the spherical surface using DiNe filter the DiNe convolution, as shown in the top row of Fig. \ref{fig2}A. With the designed filter, DiNe convolution can be formulated as a simple filter weighting process. For each vertex $v$ on a standard spherical surface with $N$ vertices, at a certain convolution layer with input feature channel number $D$ and output feature channel number $F$, the feature data $ {\bf{I_v}} (7\times D)$ from the direct neighbors are first extracted and reshaped into a row vector $ {\bf I_v'} (1 \times 7D)$. Then, iterating over all $N$ vertices, we can obtain the full-node filter matrix $ {\bf I} (N\times 7D)$. By multiplying ${\bf I}$ with the convolution layer’s filter weight ${\bf W} (7D \times F)$, the output surface feature map ${\bf O} (N \times F)$ with $F$ channels can be obtained.
\begin{figure}[t]
	\centering
	\includegraphics[width=0.7\textwidth]{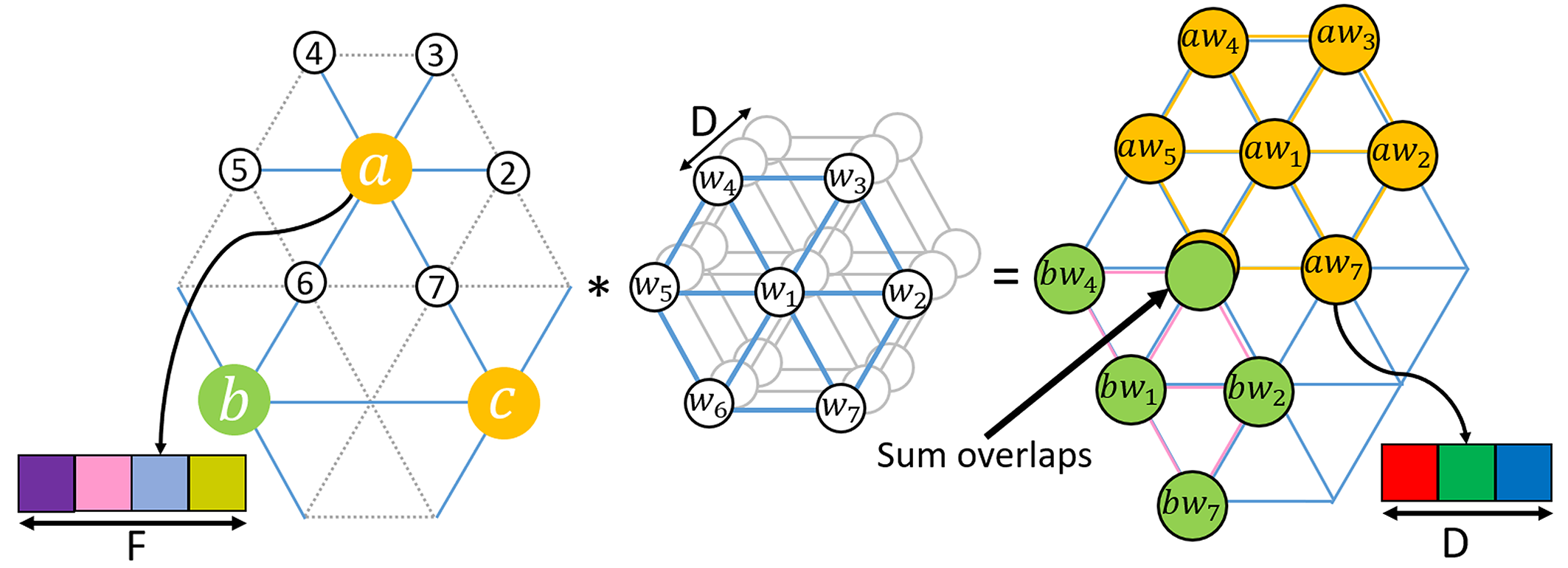}
	\caption{Illustration of transposed convolution on the spherical surface.} \label{fig3} 
\end{figure} 

The pooling operation on the spherical surface is performed in a reverse order of the icosahedron expansion process. As shown in Fig. \ref{fig2}B, in a pooling layer, for each center vertex $v$, all feature data $ {\bf I_v} (7 \times F)$ aggregated from itself and its neighbors are averaged or maximized, and then a refined feature ${\bf I_v'} (1 \times F)$ can be obtained. Meanwhile, the number of vertices is decreased from $N$ to $(N+6)/4$.

\subsection{Transposed Convolution on Spherical Surface}
Transposed convolution is also known as fractionally-strided convolution, deconvolution or up-convolution in U-Net \cite{p4}. It has been widely used for its learnable parameters in conventional CNN, especially in semantic segmentation. From the perspective of image transformation, transposed convolution first restores pixels around every center pixel by sliding-window filtering over all original pixels, and then sums where output overlaps. 

Inspired by this concept, for a spherical surface with the original feature map $ {\bf I}$($N\times D$, where $N$ denotes the number of vertices, and $D$ denotes the number of feature channels) and the pooled feature map ${\bf O}$($N' \times F$, $N'=(N+6)/4$), we can restore ${\bf I}$ by first using DiNe filter to do transposed convolution with every vertex on the pooled surface ${\bf O}$ and then summing overlap vertices (see Fig. \ref{fig3}). 

\subsection{Spherical U-Net Architecture}
With our defined operations for spherical surface convolution, pooling, and transposed convolution, the proposed Spherical U-Net architecture is illustrated in Fig. \ref{fig1}. It has an encoder path and a decoder path, each with five resolution steps, indexed by $i$, $i=1,2,\dots,5$. Different from the standard U-Net, we replace all 3$\times$3 convolution with our DiNe convolution, 2$\times$2 up-convolution with our surface transposed convolution, and 2$\times$2 max pooling with our surface max/mean pooling. In addition to the standard U-Net, before each convolution layer’s rectified linear units (ReLU) activation function, a batch normalization layer is added. At the final layer, 1$\times$1 convolution is replaced by vertex-wise filter weighting to map $C_1$-component feature vector in the second-to-last layer to the desired $C_{out}$ at the last layer. We simply double the number of feature channels after each surface pooling layer and halve the number of feature channels at each transposed convolution layer. That makes $C_{i+1}=C_i\times 2$ and $N_{i+1}=(N_i+6)/4$, as shown in Fig. \ref{fig1}. Note that we do not need any tiling strategy in the original U-Net \cite{p4} to allow a seamless output map, because all the data flow in our network is on a closed spherical surface.

\section{Experiments}
To validate the proposed Spherical U-Net on cortical surfaces, we conducted experiments on two challenging tasks in infant brain MRI studies: cortical surface parcellation and cortical attribute map development prediction. Both tasks are of great neuroscientific and clinical importance and are suffering from designing of hand-crafted features and heavy computational burden. We show that our task-agnostic and feature-agnostic Spherical U-Net is still capable of learning useful features for these different tasks.

\subsection{Infant Cortical Surface Parcellation}

\subsubsection{Dataset and Image Processing.} 
We used an infant brain MRI dataset with 90 term-born neonates. All images were processed using an infant-specific computational pipeline \cite{p15}. All cortical surfaces were mapped onto the spherical space and further resampled. Each vertex on the cortical surface was coded with 3 shape attributes, i.e., the mean curvature, sulcal depth, and average convexity. The target is to parcellate these vertices into 36 cortical regions for each hemisphere. A 3-fold cross-validation was adopted and Dice ratio was used to measure the overlap between the manual parcellation and the automatic parcellation.

\subsubsection{Architectures.} 
We used the Spherical U-Net architecture as shown in Fig. \ref{fig1}. We set $C_{in}$ as 3 for the 3 shape attributes, $C_{out}$ as 36 for the 36 labels of ROIs, $N_1$ as 10,242, and $C_1$ as 64. For comparison, we created the following architecture variants.

As RePa convolution is very memory-intensive for a full Spherical U-Net experiment, we created a smaller variant \textit{U-Net18-RePa}. It is different from the Spherical U-Net in three points: 1) It only consists of three pooling and three transposed convolution layers, thus containing only 18 convolution layers; 2) It replaces all DiNe convolution with RePa convolution; 3) The feature number is halved at each corresponding layer. Meanwhile, for a fair comparison, we created a \textit{U-Net18-DiNe} by replacing all RePa convolution with DiNe convolution in U-Net18-RePa. \textit{Naive-DiNe} is a baseline architecture with 16 DiNe convolution blocks (DiNe (64 convolution filters), BN, ReLU) and without any pooling and upsampling layers.

\begin{figure}[t]
	\centering
	\includegraphics[width=0.75\textwidth]{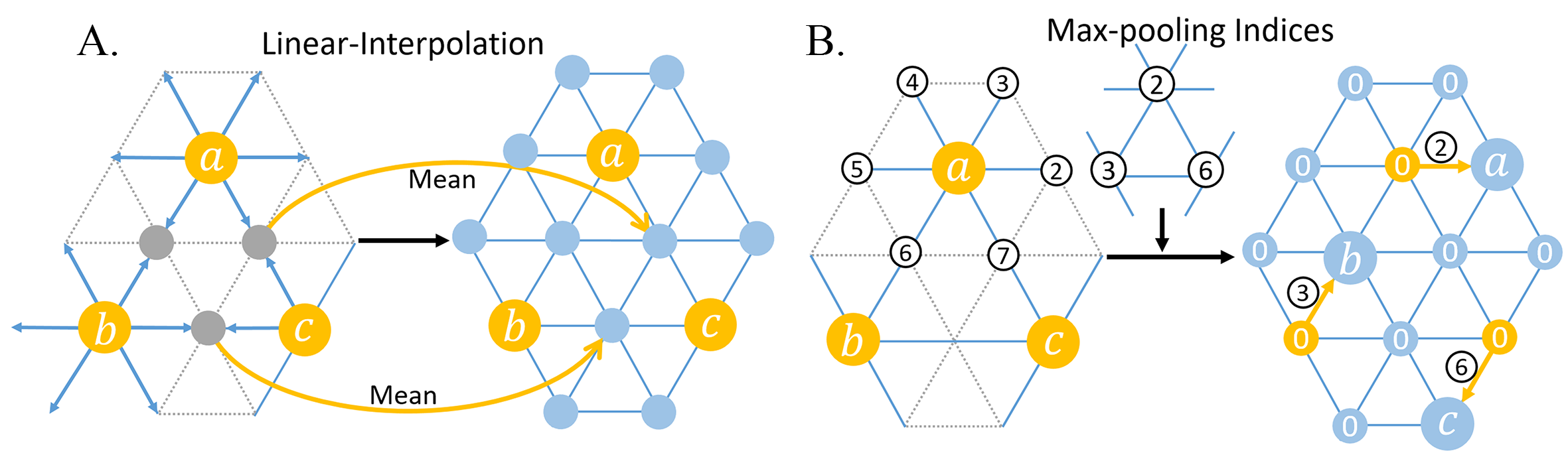}
	\caption{Illustration of Linear-Interpolation and Max-pooling Indices upsampling methods.} \label{fig4} 
\end{figure}

In addition to the above variants, we also studied upsampling using Linear-Interpolation (\textit{SegNet-Inter}) and Max-pooling Indices (\textit{SegNet-Basic}). As shown in Fig. \ref{fig4}A, for each new vertex generated from the edge’s center, its feature is linearly interpolated by the two parent vertices of this edge using Linear-Interpolation. Max-pooling Indices, introduced by SegNet \cite{p16}, uses the memorized pooling indices computed in the max-pooling step of the encoder to perform non-linear upsampling at the corresponding decoder. We accommodated this method to the spherical surface mesh as shown in Fig. \ref{fig4}B. For example, max-pooling indices 2, 3, and 6 are first stored for vertices a, b, and c, respectively. Then at the corresponding upsampling layer, the 2-nd neighbor of a, 3-rd neighbor of b, and 6-th neighbor of c are restored with a, b and c's value, respectively, and other vertices are set as 0. Therefore, SegNet-Basic and SegNet-Inter require no learning for upsampling and thus are created in a SegNet style. They are different from our Spherical U-Net in two aspects: 1) There is no copy and concatenation path in both models; 2) For up-sampling, SegNet-Basic uses Max-pooling Indices and SegNet-Inter uses Linear-Interpolation.

\subsubsection{Training.}
We trained all the variants using mini-batch stochastic gradient descent (SGD) with initial learning rate 0.1 and momentum 0.99 with weight decay 0.0001. Given different network architectures, we used a self-adaption strategy for updating learning rate, which reduces the learning rate by a factor of 5 once training Dice stagnates for 2 epochs. This strategy allowed us to achieve a gain in Dice ratio around 3\% for most architectures. We used the cross-entropy loss as the objective function for training. The other hyper-parameters were empirically set by babysitting the training process. We also augmented the training data by randomly rotating each sphere to generate more training samples.
\subsubsection{Results.}
We report the means and standard deviations of Dice ratios based on the 3-fold cross-validation, as well as the number of parameters, memory storage and time for one inference on a NVIDIA Geforce GTX1060 GPU, in Table \ref{tab1}. As we can see, our Spherical U-Net architectures consistently achieve better results than other methods, with the highest Dice ratio 88.87\%. It is also obvious that RePa convolution is more time-consuming and memory-intensive, while our DiNe convolution is 7 times faster than RePa, 5 times smaller on memory storage and 3 times lighter on model size. Moreover, it outperforms the state-of-the-art DCNN method \cite{p9} using a deep classification CNN architecture based on the projected patches on the tangent space. They reported the DCNN without graph cuts achieves the average Dice ratio 86.18\%, and the DCNN with graph cuts achieves the average Dice ratio 87.06\%. As in \cite{p9}, we also incorporated the graph cuts method for post-processing the output of our spherical U-Net, but this step has no further improvement in quantitative results. This may indicate that our Spherical U-Net is capable of learning spatially-consistent information in an end-to-end way without post-processing.
\begin{table}[t]
	\caption{Comparison of different architectures for cortical surface parcellation. The p-values are the results of paired t-test vs. Spherical U-Net.}\label{tab1}
	\begin{tabular}{L{0.29\textwidth} C{0.14\textwidth}  C{0.11\textwidth}  C{0.12\textwidth} C{0.15\textwidth} C{0.14\textwidth}}
		\hline
		Architectures & Parameters (MB) & Storage (MB) & Inference time (ms) &  Dice (\%) &  p-value \\
		\hline
		\multicolumn{6}{c}{\textit{Learning for upsampling}} \\
		\hline
		Spherical U-Net        & 26.9 & 1635  & 18.3 & \textbf{88.87$\pm$2.43} &	N.A.  \\
		Spherical U-Net18-DiNe & \textbf{1.7 } & \textbf{955}   & \textbf{8.9}  & 88.05$\pm$2.46 &	$ 1.96 \times 10^{-3} $\\
		Spherical U-Net18-RePa & 5.2  & 5047  & 64.5 & 88.28$\pm$2.50 & $ 4.92 \times 10^{-2} $ \\
		\hline
		\multicolumn{6}{c}{\textit{No learning for upsampling}} \\
		\hline
		Spherical Naive-DiNe   & 0.4  & 1499 & 15.8  & 81.74$\pm$4.96 & $ 4.87 \times 10^{-11} $  \\
		Spherical SegNet-Basic & 14.5 & 1341 & 113.5 & 78.31$\pm$4.62 & $ 5.87 \times 10^{-18}  $\\
		Spherical SegNet-Inter & 22.0 & 1533 & 20.1  & 75.12$\pm$8.39 & $ 4.57 \times 10^{-11}  $\\
		\hline
	\end{tabular}
\end{table}
\begin{figure}[t]
	\centering
	\includegraphics[width=0.58\textwidth]{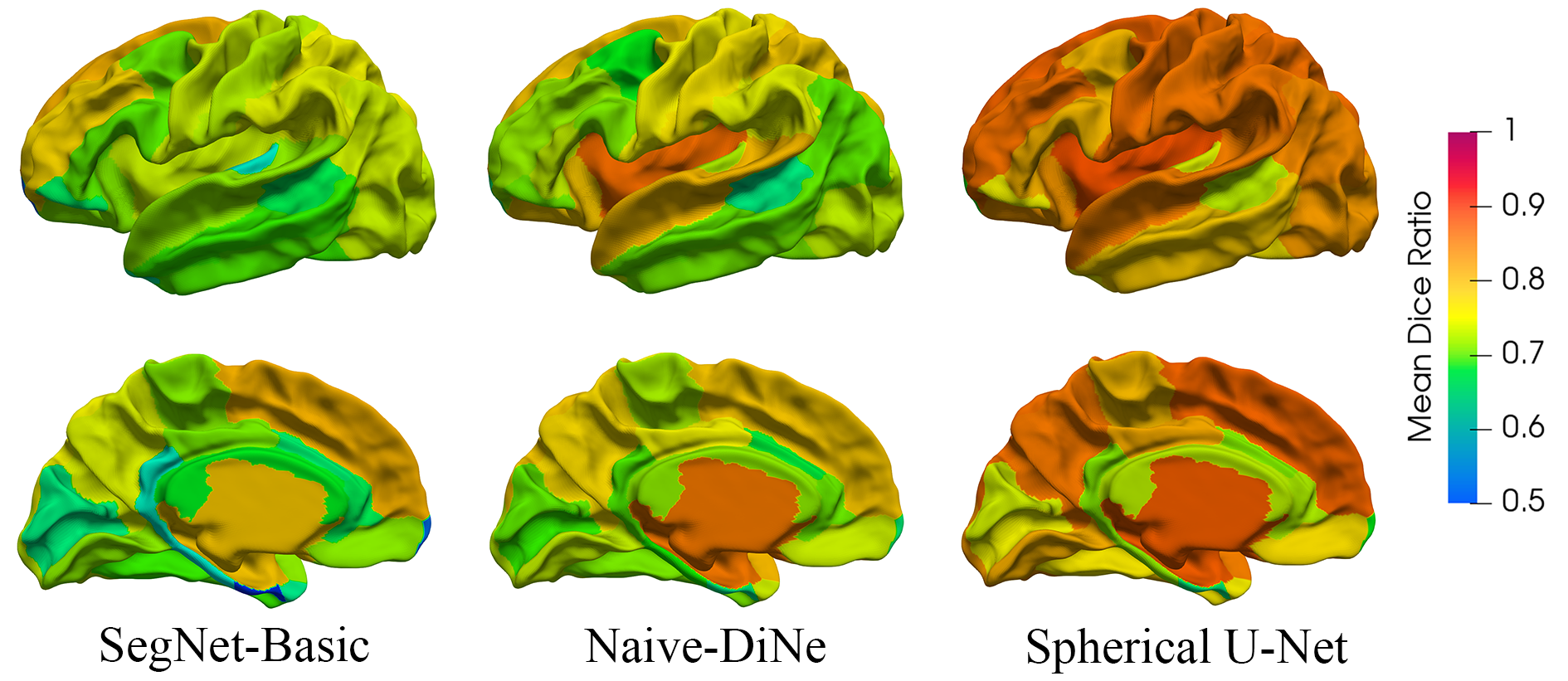}
	\caption{Average Dice ratio of cortical parcellation results for each ROI by different methods.} \label{fig5} 
\end{figure} 
\begin{figure}[t]
	\centering
	\includegraphics[width=0.69\textwidth]{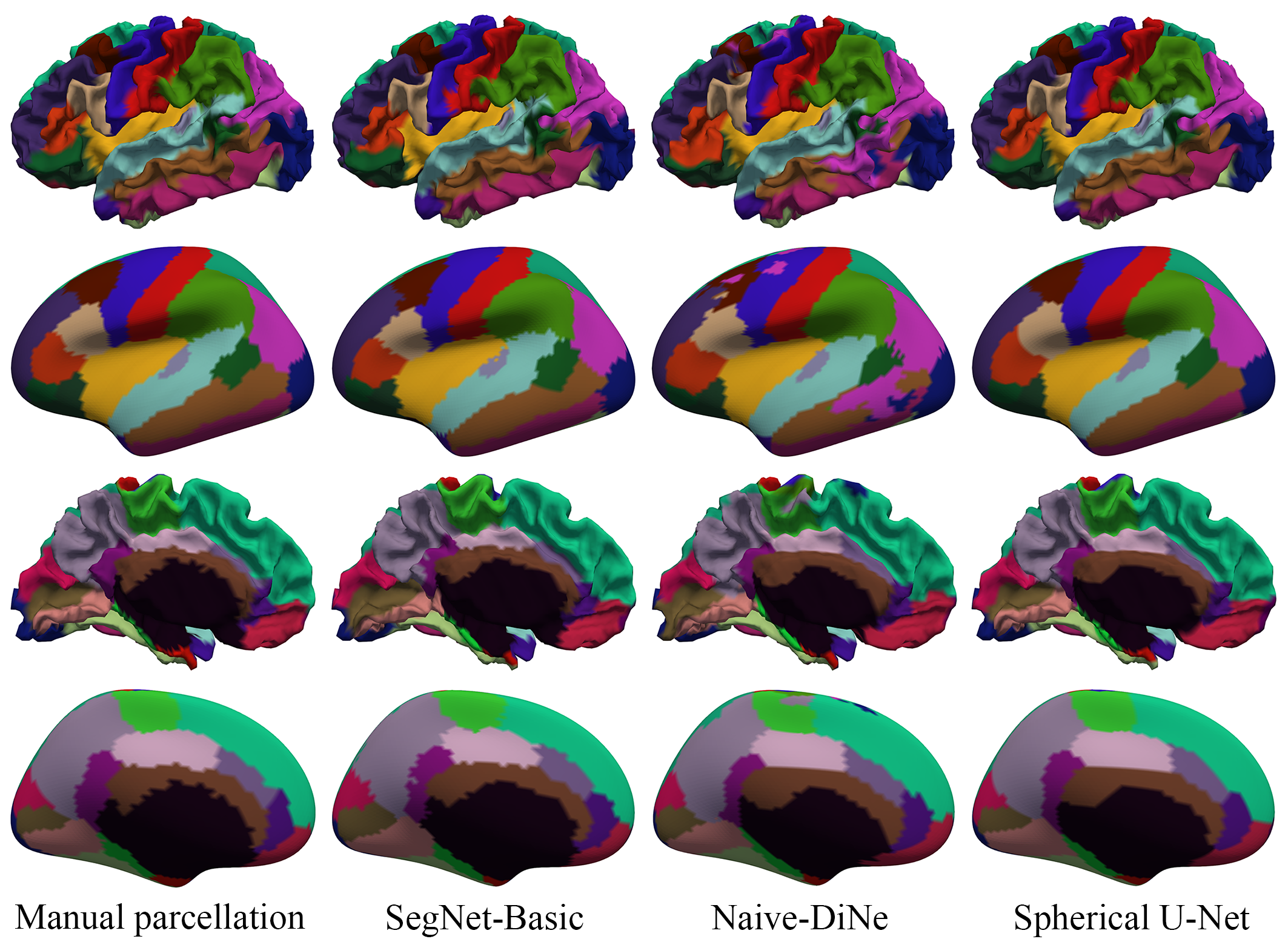}
	\caption{Visual comparison of cortical parcellation results of a randomly selected infant using different methods.} \label{fig6} 
\end{figure}

Fig. \ref{fig5} provides a comparison of average Dice ratio of each ROI using different methods. We can see that the Spherical U-Net achieves consistent higher Dice ratio in almost all ROIs. Fig. \ref{fig6} provides a visual comparison between parcellation results on a randomly selected infant by different methods. We can see that our spherical U-Net shows high consistency with the manual parcellation and has no isolated noisy labels. 
\subsection{Infant Cortical Attribute Map Development Prediction}
We have also applied our Spherical U-Net to the prediction of cortical surface attribute maps of 1-year-old brain from the corresponding 0-year-old brain using 370 infants, all with longitudinal 0-year-old and 1-year-old brain MRI data. All infant MR images were processed using an infant-specific computational pipeline for cortical surface reconstruction \cite{p15}. All cortical surfaces were mapped onto the spherical space, nonlinearly aligned, and further resampled. Following the experimental configuration in Meng et al. \cite{p17}, we used the sulcal depth and cortical thickness maps at birth to predict the cortical thickness map at 1 year of age. The reason to choose the cortical thickness map as the prediction target for validating our method is that cortical thickness has dynamic, region-specific and subject-specific development and is highly related to future cognitive outcomes \cite{p19}. To have a robust prediction for the cortical thickness, we introduced the sulcal depth as an additional channel for leveraging the relationship between sulcal depth and cortical thickness maps \cite{p23,p20}.  
\subsubsection{Evaluation Metrics.}
The evaluation metrics we adopted for the prediction performance are mean absolute error (MAE) and mean relative error (MRE) under a 5-fold cross-validation. The 5-fold cross-validation uses 60\% data for training, 20\% data for validation, and 20\% data for testing at each fold.
\subsubsection{Spherical U-Net and Hyper-parameters.}
Here we still consider a basic and simple architecture and training strategy to validate the effectiveness of our Spherical U-Net. We used the Spherical U-Net architecture as shown in Fig. \ref{fig1}, with $C_{in}=2$ (representing sulcal depth and cortical thickness channels at birth), $C_{out}$=1 (representing cortical thickness at 1 year of age), $C_1$=64, and $N_1$=40,962. We trained the Spherical U-Net using Adam optimization algorithm and L1 loss. We used an initial learning rate 0.0001 and reduced it by 10 every 3 epochs. The whole training process had 15 epochs and lasted for 30 minutes in a NVIDIA Geforce GTX1080 GPU.

\subsubsection{Comparison with Feature-based Approaches.}
For the feature-based approaches, we extracted 102 features for each vertex on 0-year-old cortical surface. Same as in Meng et al. \cite{p17}, the 1st and 2nd features are sulcal depth and cortical thickness, respectively, providing local information at each vertex. The 3rd to 102nd features are contextual features, providing rich neighboring information for each vertex, which are composed of 50 Haar-like features of sulcal depth and 50 Haar-like features of cortical thickness. The Haar-like features were extracted using the method and hyper-parameters in \cite{p17}. 

We then trained the following machine learning algorithms on the 102 features in a vertex-wise manner: Linear Regression, Polynomial Regression, and Random Forest \cite{p17}. Linear Regression assumes that cortical thickness at each vertex is linearly increased, and Polynomial Regression assumes that it has a two-order polynomial relationship with the age. Random Forest is an effective method for high dimensional data analysis, which has shown the state-of-the-art performance for cortical thickness prediction \cite{p17}. Herein, each above algorithm would generate 40,962 models, each for predicting the thickness of a certain vertex at 1-year-old, while our Spherical U-Net just generates one model for all vertices. All the machine learning algorithms were trained on a campus-wide cluster and the training process all lasted an extremely long time (2-3 days).
\begin{table}[t]
	\caption{ 5-fold cross-validated cortical thickness prediction performance in terms of MAE and MRE using different methods with standard deviations. The p-values are the results of paired t-test vs. Spherical U-Net.}\label{tab2}
	\begin{tabular}{l C{0.21\textwidth} C{0.16\textwidth}  C{0.16\textwidth}   C{0.16\textwidth} }
		\hline
		Methods & MAE (mm) & MRE (\%) & p-value for MAE & 	p-value for MRE \\
		\hline
		Linear Regression &$ 0.3605\pm 0.0337$ & $15.01\pm 1.92 $ & $9.47\times 10^{-43}$ & $1.94\times 10^{-41}$ \\
		Polynomial Regression & $0.6068\pm0.0900 $ & $26.76\pm4.52 $ & $2.01\times10^{-43}$ & $1.21\times10^{-41} $ \\
		Random Forest & $0.2959\pm0.0382 $&$ 12.63\pm2.06$ &$ 2.52\times10^{-24} $& $ 1.80\times10^{-16} $ \\
		Spherical U-Net & \textbf{0.2812$\pm$ 0.0406}  & \textbf{12.14$\pm$2.05}  & N.A. & N.A.   \\
		\hline
	\end{tabular}
\end{table}

\subsubsection{Results.}
Table \ref{tab2} presents the 5-fold cross-validation results. Our Spherical U-Net outperforms all other machine learning algorithms both in terms of MAE and MRE. While the main competitors, Random Forest is involved with complex hand-crafted features extraction step and heavy vertex-wise computational burden, our task-agnostic and feature-agnostic Spherical U-Net still achieves better results. The Linear Regression and Polynomial Regression results reveal that the cortical thickness development is more like in a linear pattern than a polynomial pattern from birth to 1 year of age, which is consistent with the finding in \cite{p17}.

\begin{figure}[t]
	\centering
	\includegraphics[width=0.85\textwidth]{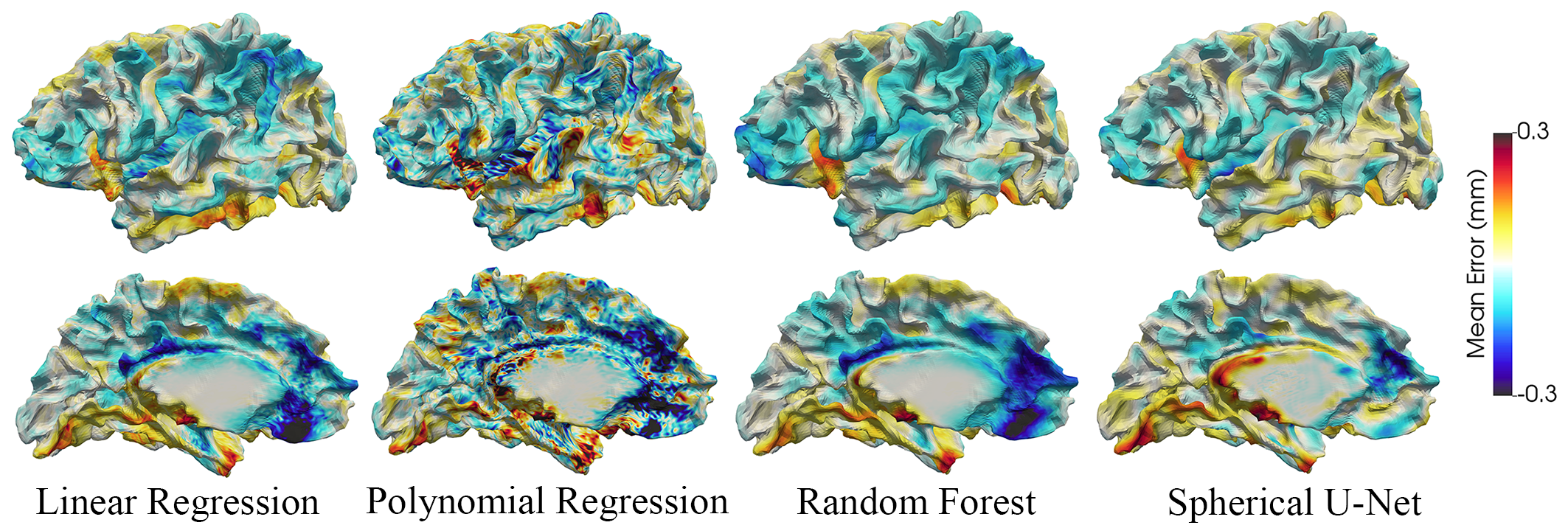}
	\caption{Visual comparison of vertex-wise average error maps using different methods.} \label{fig7} 
\end{figure}

\begin{figure}[t]
	\centering
	\includegraphics[width=\textwidth]{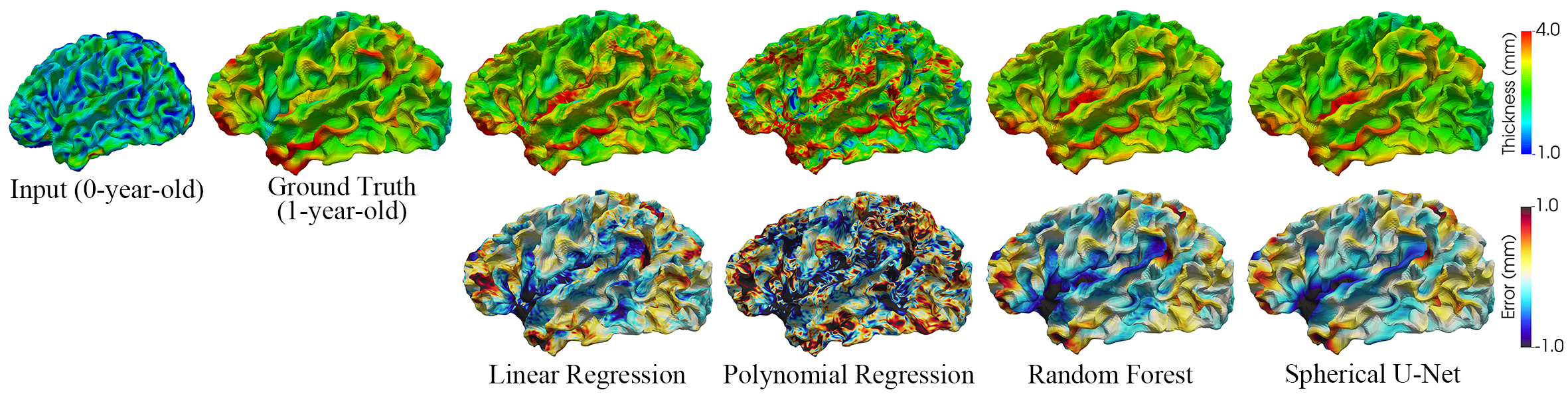}
	\caption{Prediction of the vertex-wise cortical thickness map (mm) of a randomly selected infant by different methods. The first row shows the input at 0-year-old, ground truth, and the predicted cortical thickness maps by different methods. The second row shows the error maps (mm).} \label{fig8} 
\end{figure}

Fig. \ref{fig7} shows a visual comparison on the vertex-wise mean error map between the ground truth at 1 year of age and predicted cortical thickness based on 0-year-old data using different methods. We can see that the Spherical U-Net obtains smoother and smaller mean errors than other methods. Fig. \ref{fig8} provides the vertex-wise predictions of a randomly selected infant using different methods and their corresponding error maps. As we can see, the Spherical U-Net predicts the cortical thickness map more precisely than other methods. 

\section{Conclusion}
In this paper, we propose the DiNe filter on spherical space for developing corresponding operations for constructing the Spherical CNNs. The DiNe filter has a natural and intuitive definition, making it interpretable for recognizing patterns on spherical surface. We then extend the conventional U-Net to the Spherical U-Net by deploying respective surface convolution, pooling, and transposed convolution layers. Furthermore, we have shown that the Spherical U-Net is computationally efficient and capable of learning useful features for different tasks, including cortical surface parcellation and cortical attribute map development prediction. The experimental results on these two challenging tasks confirm the robustness, efficiency and accuracy of the Spherical U-Net both visually and quantitatively. In the future, we will extensively test our Spherical U-Net on other cortical/subcortical surface tasks and also make it publicly available. 

\subsubsection{Acknowledgements.}
This work was partially supported by NIH grants (MH107815, MH108914, MH109773, MH116225, and MH117943) and China Scholarship Council.

%
% ---- Bibliography ----
%
% BibTeX users should specify bibliography style 'splncs04'.
% References will then be sorted and formatted in the correct style.
%
\bibliographystyle{splncs04}
\bibliography{references.bib}

%
%\begin{thebibliography}{8}
	
%\bibitem{ref_article1}
%Author, F.: Article title. Journal \textbf{2}(5), 99--110 (2016)

%\bibitem{ref_lncs1}
%Author, F., Author, S.: Title of a proceedings paper. In: Editor,
%F., Editor, S. (eds.) CONFERENCE 2016, LNCS, vol. 9999, pp. 1--13.
%Springer, Heidelberg (2016). \doi{10.10007/1234567890}
%
%\bibitem{ref_book1}
%Author, F., Author, S., Author, T.: Book title. 2nd edn. Publisher,
%Location (1999)
%
%\bibitem{ref_proc1}
%Author, A.-B.: Contribution title. In: 9th International Proceedings
%on Proceedings, pp. 1--2. Publisher, Location (2010)
%
%\bibitem{ref_url1}
%LNCS Homepage, \url{http://www.springer.com/lncs}. Last accessed 4
%Oct 2017
%\end{thebibliography}
\end{document}